\documentclass[10pt,journal,compsoc]{IEEEtran}
\usepackage{amsmath,amsfonts}
\usepackage[noend]{algpseudocode}
\usepackage{algorithm}
\usepackage{algorithmicx}
\usepackage{array}
\usepackage[caption=false,font=normalsize,labelfont=sf,textfont=sf]{subfig}
\usepackage{textcomp}
\usepackage{stfloats}
\usepackage{url}
\usepackage{verbatim}
\usepackage{graphicx}
\usepackage{cite}
\usepackage{multirow}
\usepackage{booktabs}
\usepackage{pgfplots} 
\usepgfplotslibrary{polar}
\usepackage{subcaption}
\begin{document}

\title{
EEG-Xplain: Decoding Neural Black-Boxes of EEG Foundation Models
}

\author{Hansong Ma, 
        Junxiao Wang
\thanks{
Corresponding author: Junxiao Wang.}
\thanks{Hansong Ma and Junxiao Wang are with Guangzhou University, China. E-mail: hansongma@e.gzhu.edu.cn, junxiao.wang@gzhu.edu.cn}
}

\markboth{Journal of \LaTeX\ Class Files,~Vol.~14, No.~8, August~2021}%
{Shell \MakeLowercase{\textit{et al.}}: A Sample Article Using IEEEtran.cls for IEEE Journals}

\IEEEpubid{0000--0000/00\$00.00~\copyright~2021 IEEE}

\IEEEtitleabstractindextext{%
\begin{abstract}
EEG foundation models such as BIOT, LaBraM, and EEGMamba have achieved remarkable performance in neural signal decoding, but their black-box nature limits clinical trust and neuroscientific validation. We propose a unified attribution framework for interpreting EEG foundation models across heterogeneous architectures. The framework integrates gradient-, perturbation-, and activation-based explanation methods to analyze model behavior in spatial, temporal, and frequency dimensions. Spatially, it identifies critical EEG channels and visualizes their distributions using topographic maps. Temporally, it highlights decision-relevant signal segments through attribution heatmaps. In the frequency domain, it quantifies the contributions of canonical EEG rhythms via spectral perturbation analysis. To assess explanation reliability, we introduce a population-level evaluation combining Area Over the Perturbation Curve (AOPC) and cross-method consistency analysis. The framework further leverages Large Language Models (LLMs) to transform structured attribution outputs into natural-language reports, bridging low-level neural representations and high-level semantic reasoning. Experiments on benchmark datasets, including Mumtaz2016 and TUAB, demonstrate that the generated explanations are consistent with established neurophysiological markers, validating meaningful neural representations while exposing potential dependencies on artifacts and spurious patterns. The proposed framework provides a standardized approach for evaluating the interpretability, reliability, and physiological plausibility of EEG foundation models.

\end{abstract}

}

\maketitle

\section{Introduction}
\textbf{Background.} 
Electroencephalography (EEG), serving as a core non-invasive technique for monitoring the dynamic electrophysiological activity of the human brain, holds irreplaceable value in the fields of neuroscience research, clinical auxiliary diagnosis, and Brain-Computer Interfaces (BCI). Early deep learning research primarily focused on lightweight architectures designed for specific tasks, such as EEGNet~\cite{Lawhern2016EEGNetAC}, proposed by Lawhern et al., and other variants based on Convolutional Neural Networks (CNNs). While these models successfully captured transient spatio-temporal patterns through local convolutional operators, their representational capacity was constrained by the scale of labeled data and the task-specific nature of their design.
In recent years, the field of EEG decoding has undergone a significant paradigm shift, entering the era of Foundation Models. Models exemplified by BIOT~\cite{yang2023biot}, EEGPT~\cite{wang2024eegpt}(based on the Transformer architecture), and EEGMamba~\cite{gui2024eegmamba} (based on selective state-space models) have shattered performance records across various clinical benchmark tasks, such as the Mumtaz2016 depression triage task  and the TUAB clinical abnormality detection task . These achievements were realized through large-scale self-supervised pre-training (e.g., masked autoencoding or contrastive learning) conducted on tens of thousands of hours of unlabeled EEG data. These models have demonstrated that large-scale representation learning can capture deeper, latent neurodynamic features than traditional architectures, possessing immense potential for generalization across different datasets.

\textbf{Motivation.}
Despite the breakthroughs achieved by foundation models in predictive accuracy, their massive parameter counts and complex internal logic (such as global attention mechanisms or linear state transitions) have resulted in decision-making processes that exhibit severe ``black-box'' characteristics~\cite{kuruppu2026eeg}. This lack of transparency creates a significant trust gap in high-stakes application scenarios, such as medical diagnosis. Currently, research into the interpretability of EEG foundation models still faces the following three core  bottlenecks:
\textbf{First, there exists a challenging problem regarding the misalignment of latent space representations.} EEG foundation models (such as LaBraM) map raw signals into a discrete latent space using a Neural Tokenizer. This implies that the models no longer process physical waveforms directly, but rather operate on high-dimensional combinations of symbolic tokens. The core challenge lies in this: although the model achieves accurate classification, there is a lack of a ``Mapping Protocol'' to bridge the features learned internally by the model with established neurophysiological principles (such as specific patterns of rhythmic evolution). Existing attribution methods can only provide local sensitivity maps; they are unable to \textbf{distill} the model's decision paths within the latent space into clinically interpretable evidence through dimensionality reduction.

\textbf{Secondly, the risk of spurious causality is a pervasive issue within the learning pathways of foundation models.} Foundation models, trained on massive and heterogeneous datasets, are highly prone to acquiring statistical shortcuts that span across different clinical centers and recording devices~\cite{bommasani2021opportunities,geirhos2020shortcut}. A critical challenge lies in determining whether an EEG foundation model bases its judgments on genuine physiological significance; we need a framework capable of simultaneously identifying the regions the model deems critical and evaluating whether the model made the correct decision based on the right reasons. This is particularly critical in high-stakes scenarios such as medical diagnosis, as erroneous decisions could hinder the clinical deployment of the model. For instance, if a model achieves accurate classification solely by detecting electrode impedance imbalance( a type of physical artifact), it would lack clinical credibility~\cite{ghassemi2021false}.

\textbf{Furthermore, there exists a distinct logical disconnect between perceptual maps and semantic reasoning.} Fundamentally, current Explainable AI (XAI) results remain confined to the realm of visual features rather than constituting knowledge-based conclusions. The key requirement is that neuroscientists and clinicians need a summary of the causal chain; for example, a diagnosis of a high level of attention is inferred from the suppression of alpha waves in the occipital lobe. Existing attribution algorithms fail to bridge the gap between gradient-based weights and symbolic reasoning, thereby creating a significant barrier to effective interaction between AI-driven decision-making and the clinical logic employed by human experts.

\textbf{Our contributions.}
Addressing the aforementioned challenges, this paper proposes a unified attribution framework that integrates heterogeneous foundation models, multi-faceted attribution algorithms, and Large Language Models. The objective of this framework is to distill high-dimensional latent representations into low-dimensional physical evidence, ultimately translating this evidence into semantic reports that are readily interpretable by clinical experts. The main contributions of this study are summarized as follows:
\begin{itemize}
\item \textbf{Integration of a Unified Cross-Architecture Attribution Framework}. We have built a general-purpose analysis platform compatible with various advanced foundation model architectures (such as BIOT, LaBraM, EEGPT, and EEGMamba) and multiple types of post-hoc attribution algorithms, including input gradient-based, perturbation-based, and feature activation-based methods. Through this platform, we are able to systematically compare the impact of different pre-training strategies on feature capture preferences (e.g., spatial localization accuracy and time-frequency sensitivity).

\item \textbf{Quantified Fidelity Analysis and Model Diagnosis Paradigm.} 
We introduced a population-level fidelity analysis paradigm based on Progressive Occlusion and the Area Over Perturbation Curve (AOPC)~\cite{samek2016evaluating}. By quantitatively comparing the rate and significance of the decline in model confidence following the removal of the top-$k$ core features, we validate the causal consistency between attributional evidence and the model's underlying logic. This process reveals the neurophysiologically aligned features learned by the model, as well as any potential dependencies on erroneous patterns.

\item \textbf{Preliminary Exploration of Neuro-Symbolic Semantic Reasoning Pathways.} 
We innovatively utilized the structured evidence (in JSON or image format) generated by multi-dimensional attribution as prompts for a Large Language Model (LLM). Leveraging the LLM's medical knowledge base, we achieved an automated translation from low-level physical representations to high-level clinical semantics, thereby providing neuroscientists with intuitive and transparent decision-making reports .
\end{itemize}

Our code is available on Github\footnote{https://github.com/gzhu-hcai/EEG-Xplain}. 

\section{Related Work}
\subsection{Evolution of EEG Foundation Models}
Early research on automated Electroencephalography (EEG) decoding primarily relied on supervised learning architectures designed for specific tasks. Convolutional Neural Networks (CNNs), such as EEGNet~\cite{lawhern2018eegnet} (proposed by Lawhern et al.) and DeepConvNet~\cite{schirrmeister2017deep} (proposed by Schirrmeister et al.), achieve efficient extraction of spatiotemporal features by introducing local constraints across spatial and temporal dimensions. However, these models were typically constrained by the limited scale of specific datasets, making it difficult to overcome the challenges posed by the extreme inter-individual variability and non-stationarity inherent in EEG signals.

In recent years, inspired by the paradigm of large-scale pre-training, EEG Foundation Models (FMs) have emerged at the forefront of the field. BENDR~\cite{kostas2021bendr}, proposed by Kostas et al., marked the inception of large-scale self-supervised pre-training for EEG. Subsequently, Yang et al.~\cite{yang2023biot} introduced a channel-level spectral tokenization mechanism through BIOT, enabling a unified representation of heterogeneous lead data. Inspired by the Masked Autoencoder (MAE) framework, LaBraM~\cite{jiang2024large} utilizes a neural tokenizer to map continuous signals into a discrete latent space, thereby significantly improving the modeling of long-range neurodynamic features. Furthermore, EEGMamba~\cite{gui2024eegmamba}, based on Selective State Space Models, further optimized the computational efficiency of sequence modeling. Although Foundation Models have achieved State-of-the-Art (SOTA) performance across various tasks, their massive parameter counts and patch-based discretization logic endow their internal knowledge representations with highly ``black-box'' characteristics; consequently, the alignment between their decision-making logic and established neurophysiological principles remains an area requiring systematic investigation.
\subsection{Post-hoc Attribution and XAI Applications in EEG}
Applying post-hoc attribution algorithms to EEG models to identify decision-relevant features across spatial and temporal dimensions has become a focal point of research. Sturm et al.~\cite{sturm2016interpretable} were the first to propose integrating Deep Neural Networks (DNNs) with Layer-wise Relevance Propagation (LRP) for EEG data analysis, successfully revealing feature patterns that align with neurophysiological principles. Schirrmeister et al.~\cite{schirrmeister2017deep}, through systematic feature visualization and perturbation analysis, revealed the heavy reliance of CNNs on specific frequency bands (such as Alpha and Beta rhythms) during the decoding process. This finding established the efficacy of deep learning models in extracting features within the frequency domain.

With the evolution of algorithms, SHAP (Shapley Additive Explanations) as proposed by Lundberg et al.~\cite{lundberg2017unified} has been widely adopted to interpret complex neurophysiological decision-making processes, largely due to its robust foundation in game theory. Shawly et al.~\cite{shawly2025eeg} integrated novel attention modules into CNNs and utilized SHAP to introduce quantitative interpretability metrics, thereby refining the logic behind the extraction of EEG signal features. Islam et al.~\cite{islam2022explainable} employed the LIME algorithm to interpret the behavior of a stroke prediction model, successfully identifying biological features that contribute significantly to clinical diagnosis. In the realm of clinical applications, Khan et al.~\cite{khan2025explainable} proposed a framework for epilepsy detection that combines machine learning with XAI techniques, thereby significantly enhancing the system's credibility in supporting clinical diagnosis. Addressing emerging foundation model architectures, Madsen et al.~\cite{madsen2023concept} introduced the Concept Activation Vectors (CAV) technique into the field of EEG analysis; by leveraging external labeled data and anatomical definitions to construct conceptual representations, they successfully improved the interpretability of large-scale Transformer models such as BENDR. However, most existing studies focus on single-model designs tailored to specific tasks. They lack a standardized auditing framework capable of universally adapting to heterogeneous foundation models and translating weight distributions into physiologically meaningful evidence.

\subsection{Neuro-Symbolic Integration and Semantic Reasoning via LLMs}
Translating low-level attribution maps into high-level semantic logic is crucial to bridge the interpretability gap. With the maturation of LLMs, their potential to assist in medical decision-making and scientific logical reasoning has been widely demonstrated. A study by Singhal et al.~\cite{singhal2023large} demonstrates that LLMs possess extensive clinical medical knowledge and exhibit exceptional potential for diagnostic reasoning. In the field of EEG, Thapa et al.~\cite{kim2024eeg} introduced EEG-GPT, conducting a preliminary exploration into the feasibility of using LLMs to interpret EEG feature descriptors. A current limitation is that, while post-hoc attribution techniques can generate high-quality spatial topological maps and time-frequency evidence spectra, deriving logically consistent diagnostic conclusions from these visualizations remains a significant cognitive challenge for non-clinician experts. This paper aims to address this challenge through neuro-symbolic integration: first, relevant physical evidence (such as specific spatial polarities and time-frequency patterns) is tokenized into structured JSON data or direct input; subsequently, leveraging the medical knowledge base of LLM, this data is transformed into an expert report in natural language. This closed-loop process not only enhances the interpretability of the underlying model but also offers a novel theoretical paradigm for constructing end-to-end, auditable EEG AI systems.

\begin{figure*}[t]
\centering
\includegraphics[width=6.8in]{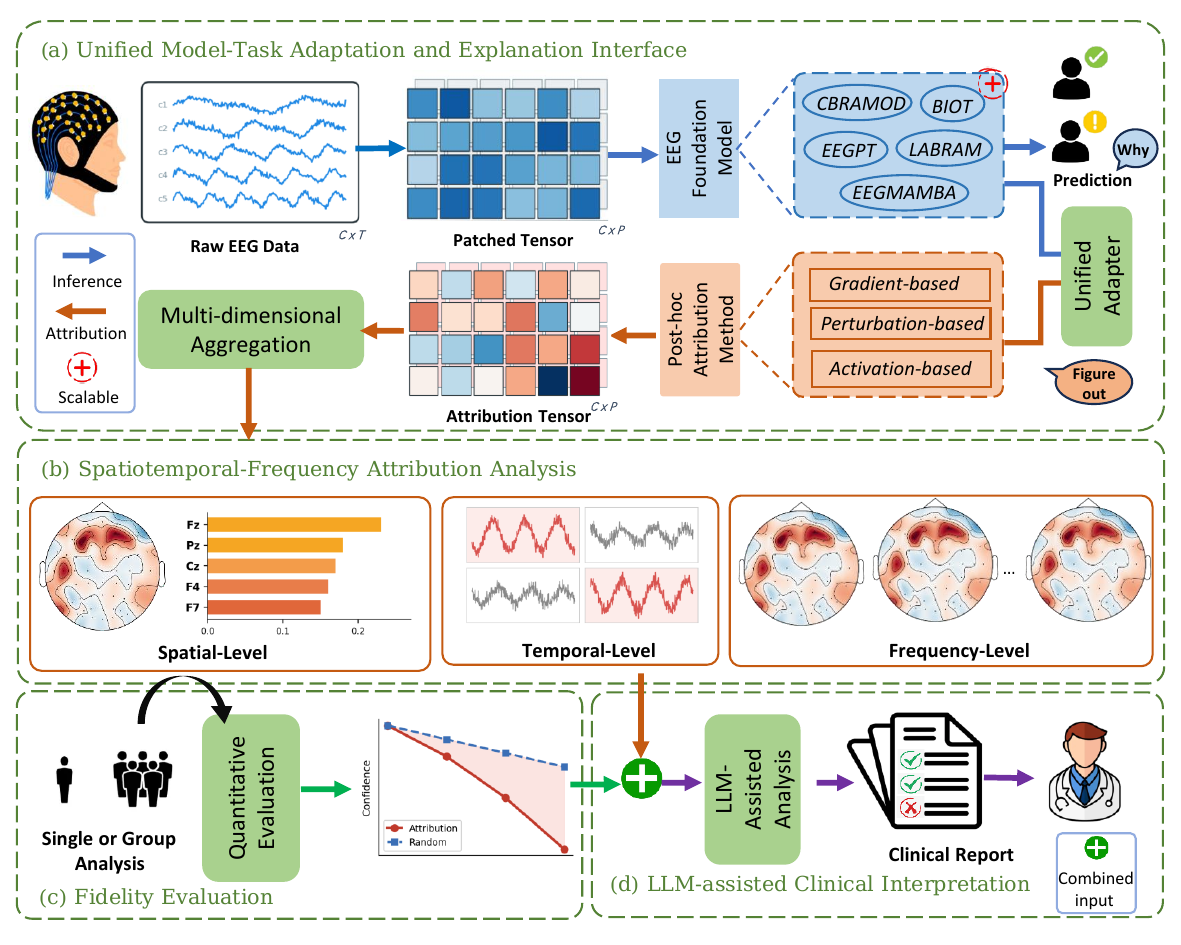}
\caption{
Overview of our proposed EEG-Xplain framework. It comprises four components: (a) a unified interface for various models and tasks to generate diverse types of model prediction attributions, with an extensible list of supported models and tasks; (b) projects attribution results onto three distinct dimensions: spatial, temporal, and frequency-band; (c) supports both single-sample and population-level analyses to verify the reliability of the attribution results; and (d) integrates the multidimensional analysis and validation results as joint inputs, leveraging the medical knowledge of large-scale models to align the attributions and analyses with physiological significance, ultimately generating a structured report for expert review.
}
\label{fig:fig2}
\hfil
\end{figure*}

\section{Method}
\subsection{Preliminaries}
\label{subsec:preliminaries}

\noindent \textbf{EEG Signal Modeling and  Characteristics of Foundation Model Architecture.} 
Let $\mathbf{X} \in \mathbb{R}^{C \times T}$ represent a multi-channel EEG recording, where $C$ is the number of electrodes and $T$ is the number of time samples. An EEG foundation model (FM) is a pre-trained backbone $f_\theta$ that maps $\mathbf{X}$ into a high-dimensional representation space $\mathbf{H}$. In downstream applications (e.g., Mumtaz2016 for depression detection or TUAB for abnormality screening), the model is typically appended with a task-specific head $h_\phi$ to produce a prediction $y = h_\phi(f_\theta(\mathbf{X}))$. For classification, $y$ denotes the class probability after softmax normalization.
Unlike traditional end-to-end CNNs, current FMs for EEG employ large-scale self-supervised pre-training and incorporate the following three key architectural features:
\begin{itemize}
    \item \textit{Tokenization and Patching:} These models utilize patch-based representations, segmenting the raw signal $X$ into patches along the time axis and learning features at the patch level.
    \item \textit{Encoder Architectures:} FMs primarily leverage Transformer encoders to capture global attention across tokens, or Selective State Space Models (SSMs) like EEGMamba to model long-range dependencies with linear complexity. 
    \item \textit{Latent Space Representation:} The decision-making process of these models relies on high-dimensional, abstract representations within the encoder's hidden layers; consequently, the specific contributions of input channels and time segments to the prediction results cannot be directly interpreted.
\end{itemize}

\noindent \textbf{Paradigms of Post-hoc Attribution.} 
To systematically audit the decision-making process of the EEG foundation models, we integrate six representative algorithms, categorized into three mathematical paradigms:

\noindent \textbf{(1) Input Gradient-based Paradigm (IG, Gradient-SHAP):} 
These methods quantify feature importance by backpropagating the model's prediction to the input space. The core idea is to measure the sensitivity of the output $f(\mathbf{X})$ via gradients. 
\begin{itemize}
    \item \textit{Integrated Gradients (IG)} calculates the integral of gradients along a straight-line path from a baseline $\mathbf{X}'$ to the input $\mathbf{X}$, ensuring the \textit{Completeness} axiom:
    \begin{equation}
        \phi_{IG}(\mathbf{X}) = (\mathbf{X} - \mathbf{X}') \times \int_{\alpha=0}^1 \frac{\partial f(\mathbf{X}' + \alpha(\mathbf{X} - \mathbf{X}'))}{\partial \mathbf{X}} d\alpha
    \end{equation}
    \item \textit{Gradient-SHAP} extends IG by sampling multiple interpolated paths between the baseline and input, averaging gradients across these samples to reduce variance and provide a smoother, more stable attribution map for high-dimensional EEG patches.
\end{itemize}

\noindent \textbf{(2) Perturbation-based Paradigm (Occlusion, SHAP, LIME):} 
These are model-agnostic approaches that treat the FM as a black box, estimating importance by observing the response to input variations.
\begin{itemize}
    \item \textit{Occlusion} systematically masks temporal patches or channels to measure the direct drop in confidence $\Delta y = f(\mathbf{X}) - f(\mathbf{X}_{masked})$.
    \item \textit{SHAP} leverages cooperative game theory to assign each feature a value representing its \textit{marginal contribution} across all possible feature combinations:    \begin{equation}
        \phi_i = \sum_{S \subseteq \mathcal{F} \setminus \{i\}} w(|S|) [f(S \cup \{i\}) - f(S)]
    \end{equation}
    where $S \subseteq \mathcal{F} \backslash \{i\}$ denotes all feature subsets excluding $i$, $w(|S|)$ is a weighting function, and $f(S \cup {i}) - f(S)$ measures the marginal contribution of adding feature $i$. We use \textit{KernelSHAP} for efficient approximation.
    \item \textit{LIME} approximates the FM locally by training an interpretable surrogate model $g$ (e.g., a linear regressor) on perturbed samples in the neighborhood of $\mathbf{X}$.
\end{itemize}

\noindent \textbf{(3) Feature Activation-based Paradigm (Grad-CAM):} 
Targeting models with convolutional or self-attention layers (e.g., EEG-Conformer), \textit{Grad-CAM} computes a coarse localization map via a weighted combination of feature maps, where the weights are derived from the globally pooled gradients of the final bottleneck layer:
\begin{equation}
    L_{Grad-CAM}^c = \text{ReLU} \left( \sum_k \alpha_k^c \mathbf{A}^k \right)
\end{equation}
where $\alpha_k^c$ denotes the importance weight of feature map $k$ for class $c$, computed via global average pooling of the gradients, and $\mathbf{A}^k$ represents the activation of the $k$-th feature map.

\subsection{System Architecture and Design Philosophy}
The proposed framework, as illustrated in Fig~\ref{fig:fig2}, aims to bridge the interpretability gap between the high-dimensional latent representations of EEG foundation models and tangible neurophysiological evidence. The core design philosophy is based on the concept of \textbf{Causal Distillation}: reducing millions of neural parameters into a few key spatial-temporal-frequency signatures that an expert can audit. The framework follows a hierarchical pipeline: 
(1) \textit{Unified Model-Task Adaptation and Explanation Interface}: An abstraction layer encapsulates model-specific differences to allow uniform processing of diverse architectures within a shared attribution space. It exposes a standardized attribution API, providing model-agnostic input-output specifications for downstream interpretability methods.

(2) \textit{Spatiotemporal-Frequency Attribution Analysis}: Building upon the unified interface, the framework integrates multiple post-hoc attribution algorithms (based on input gradients, perturbations, and feature activations) to generate attribution results in parallel across spatial, temporal, and frequency dimensions. The spatial dimension identifies top-$k$ channels and topological maps; the temporal dimension produces importance heatmaps overlaid on raw waveforms; and the frequency dimension quantifies the relative contributions of classic EEG rhythms via frequency-domain perturbations. This constructs a hierarchical chain of evidence ranging from coarse-grained brain regions to fine-grained spatio-temporal-frequency features. 
(3) \textit{Fidelity Evaluation}: To ensure reliability, the framework introduces a population-level evaluation mechanism, such as the Area Over the Perturbation Curve (AOPC), to quantify confidence decline during progressive masking. This mechanism provides a rigorous basis for validating attribution outcomes. 
(4) \textit{LLM-assisted Clinical Interpretation}: Structured attribution results (key channels, time intervals, frequency bands, and their relative contributions) and assessment metrics are injected into a LLM via structured prompts. The LLM automatically generates natural language interpretations consistent with neurophysiological contexts. By translating abstract attribution values into readable evidence statements for clinical experts, it completes the final distillation loop from model decisions to expert-reviewable diagnostic hypotheses.

\subsection{Unified Model-Task Adaptation and Explanation Interface}

Deep learning models for EEG data typically exhibit significant heterogeneity in their input representations; for instance, different models or tasks may employ varying numbers of channels, temporal segmentation strategies, positional encoding schemes, and feature extraction layer architectures. Directly adapting specific attribution methods to each model or task not only introduces immense implementation complexity but also compromises the comparability of the resulting explanations. To address this challenge, this paper unifies the modeling of task configurations, model adaptations, and attribution interfaces, thereby enabling the interpretation processes for diverse models to be executed within a single, consistent abstraction layer. Let the EEG input be denoted as
\begin{equation}
\mathbf{X}\in\mathbb{R}^{C\times T},
\end{equation}
where $C$ represents the number of channels and $T$ represents the temporal length. For models employing a block-based representation, this can be further expressed as
\begin{equation}
\mathbf{X}_p\in\mathbb{R}^{C\times P\times L},
\end{equation}
where $P$ is the number of temporal blocks and $L$ is the length of each block, satisfying $T=P\cdot L$. The core concept of unified adaptation is to define a mapping from the original input space to the model input space:
\begin{equation}
\phi:\mathbb{R}^{C\times T}\rightarrow \mathbb{R}^{C\times P\times L}.
\end{equation}
Based on this foundation, the models can be uniformly represented as
\begin{equation}
\hat{\mathbf{y}} = f(\phi(\mathbf{X})),
\end{equation}
where $f(\cdot)$ denotes the deep EEG network under interpretation. For a given sample and target class, various attribution methods are ultimately constrained to produce a unified joint attribution matrix
\begin{equation}
\mathbf{A}\in\mathbb{R}^{C\times P},
\end{equation}

Within the unified output space, spatial importance and temporal importance are derived through marginal aggregation, defined as follows:

\begin{equation}
s_c = \sum_{p=1}^{P} \mathbf{A}_{c,p}, \quad t_p = \sum_{c=1}^{C} \mathbf{A}_{c,p}, \quad c=1,\dots,C, \ p=1,\dots,P.
\end{equation}
Here, $s_c$ quantifies the overall contribution strength of channel $c$ to the current prediction, while $t_p$ characterizes the significance of information surrounding time block $p$ to the model's discriminative decision. By employing this unified aggregation scheme, comparisons and statistical analyses can be conducted across identical channel and temporal dimensions, even when the underlying attribution methods differ.

\subsection{Spatiotemporal-Frequency Attribution Analysis}

Upon obtaining the unified attribution matrix, the framework conducts an in-depth analysis of the model's discriminative basis across three dimensions: space, time, and frequency. The primary objective of this module is to transform raw, high-dimensional, and method-specific attribution results into multi-view representations with neurophysiological interpretability, thereby identifying the key brain regions, time segments, and frequency components upon which the model relies.

\subsubsection{Spatial Attribution Analysis}

\begin{figure}[htbp]
    \centering
    \includegraphics[width=0.50\textwidth]{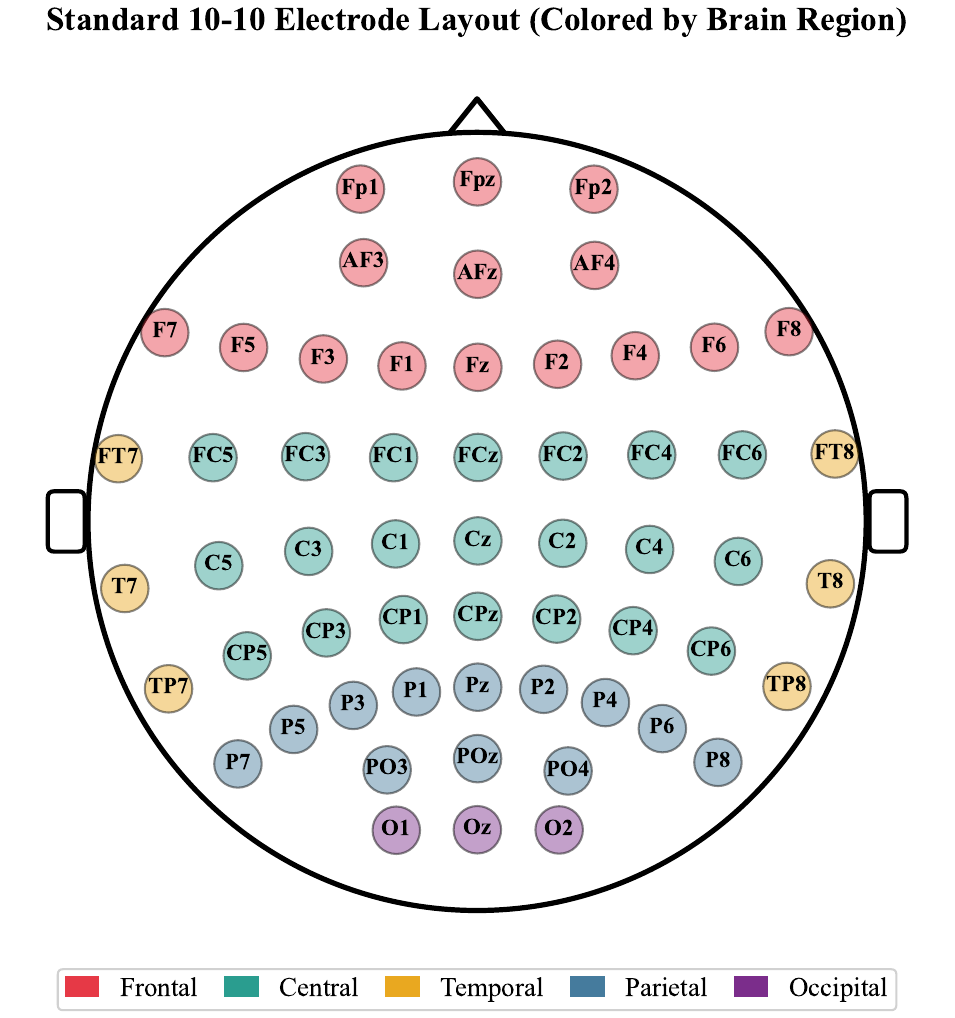}
    \caption{Standard 10-10 electrode layout colored by brain region.}
    \label{fig:electrode_map}
\end{figure}

Spatial dimension analysis is grounded in the channel importance vector $\mathbf{s}=[s_1,\dots,s_C]$, which serves to characterize the specific regions on the scalp topography to which the model directs its attention. Since EEG channels correspond to fixed electrode positions, the vector $\mathbf{s}$ can be mapped onto a standard scalp coordinate system to form a continuous topological distribution. Figure \ref{fig:electrode_map} illustrates the electrode distribution for a monopolar montage based on the standard 10-10 system; it comprises approximately 60 electrode locations, categorized into five anatomical brain regions: frontal, central, temporal, parietal, and occipital.
For group-level analysis: Assuming that there are $N$ samples in total, the importance of group-level channel  can be constructed by calculating the sample mean:
\begin{equation}
\bar{s}_c = \frac{1}{N}\sum_{n=1}^{N} s_c^{(n)}.
\end{equation}
Furthermore, these results can be aggregated at the level of anatomical regions. If $\Omega_r$ denotes the set of channels corresponding to the $r$-th brain region, the regional contribution is defined as:
\begin{equation}
S_r = \sum_{c\in\Omega_r}\bar{s}_c.
\end{equation}
This statistic can be utilized to compare the relative contributions of different brain regions to the model's decision-making process, and it also facilitates the analysis of hemispheric lateralization.

\subsubsection{Temporal Attribution Analysis}

Analysis along the temporal dimension is based on the temporal structure of the attribution matrix. 

At the sample level, each row of the attribution matrix $\mathbf{A} \in \mathbb{R}^{C \times P}$ contains $P$ patch-level attribution scores for the respective channel. Each patch $p$ corresponds to the time interval $[(p-1) \cdot S, (p-1) \cdot S + w]$ on the raw waveform time axis (where $w$ is the patch width and $S$ is the stride). By mapping these patch-level scores back to their corresponding time intervals, we enable patch-aligned temporal attribution visualization at the channel level.

At the population level, \textbf{for event-related tasks} where temporal structure is preserved across samples, we compute temporal importance by aggregating attribution values across both channels and samples:
\begin{equation}
\bar{t}_p = \frac{1}{N} \sum_{n=1}^{N} t_p^{(n)}, \quad t_p^{(n)} = \frac{1}{C} \sum_{c=1}^{C} A_{c,p}^{(n)}
\end{equation}

This yields a population-level temporal importance vector $\bar{\mathbf{t}} \in \mathbb{R}^P$, characterizing the model's average attention pattern along the time axis.

\begin{figure}[htbp]
\centering
\includegraphics[width=\linewidth]{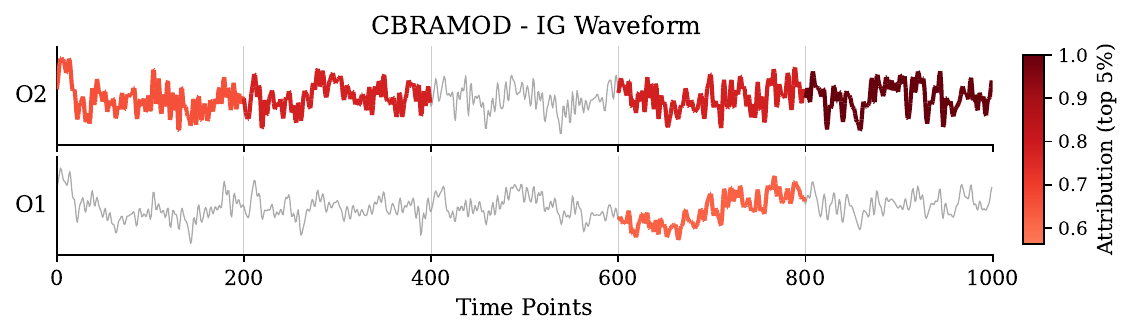}
\caption{Visualization of temporal attribution for a TP sample from a healthy subject in the Mumtaz2016 dataset.}
\label{fig:temporal_example}
\end{figure}

Temporal attribution reveals whether the model has learned to selectively focus on specific time intervals. \textbf{For event-based tasks} (e.g., epileptic discharge detection), high-contribution patches should cluster within time windows containing sharp waves or spike-and-slow-wave complexes. Conversely, \textbf{for resting-state tasks} (e.g., normal vs. abnormal classification), temporal contributions are typically more uniform, reflecting the model's reliance on global rhythmic features rather than transient events. Figure~\ref{fig:temporal_example} presents the results obtained from a sample taken in a healthy subject
from the Mumtaz2016 dataset: the top-$k$ high-contribution patches for channels O1 and O2 are evenly distributed along the time axis, indicating that the model's decision for this sample does not rely heavily on any single local time segment.

\subsubsection{Frequency Band Attribution Analysis}

Frequency band attribution aims to investigate which neural oscillation rhythms the model relies on. We define the frequency bands $\mathcal{B} = \{\delta, \theta, \alpha, \beta, \gamma\}$ (0.5--4, 4--8, 8--13, 13--30, 30--45\,Hz respectively). The framework quantifies the causal contribution of each frequency band through band ablation.

For each channel $c$ and frequency band $b$, a band-ablated baseline signal $x'_{c,b}$ is constructed: a band-stop filter is applied to channel $c$ to remove the energy components within band $b$ while preserving the remaining frequency components. Taking Integrated Gradients as an example, the attribution is computed via path integration from this baseline to the original signal:
\begin{equation}
\text{BA}_{c,b} = \sum_{t} (x_{c,t} - x'_{c,b,t}) \cdot \int_0^1 \frac{\partial F_k}{\partial x_{c,t}}(x'_{c,b} + \alpha(x_c - x'_{c,b})) \, d\alpha
\end{equation}
where $t$ indexes all time points of channel $c$. This carries a clear causal interpretation: $\text{BA}_{c,b}$ quantifies the change in the model's prediction confidence for the target class if the oscillatory activity of band $b$ were removed from channel $c$. 
By aggregating across samples, population-level band-channel attribution is obtained:
\begin{equation}
\overline{\mathbf{BA}}_{c,b} = \frac{1}{N} \sum_{n=1}^{N} \text{BA}^{(n)}_{c,b}
\end{equation}
This produces a population-level band-channel attribution matrix $\overline{\mathbf{BA}} \in \mathbb{R}^{C \times |B|}$. It allows visualization of spatial distribution patterns of various rhythms through band-specific topographic maps. For example, if the model relies on alpha rhythms in posterior regions for classification, $\overline{\mathbf{BA}}_{:,\alpha}$ should exhibit significant positive values in occipital channels (O1, O2, Oz).

\subsection{Fidelity Evaluation}

Does the importance ranking generated by the attribution method faithfully reflect the model's internal decision-making logic? The framework employs an intervention-based fidelity evaluation paradigm to verify this. The core idea is that if the attribution correctly identifies features causally contributing to the decision, then systematically removing high-attribution features should cause a more significant drop in performance than random removal.

\subsubsection{Perturbation-Based Faithfulness}

The framework conducts cumulative masking experiments on the attributed features. Let $\pi$ denote the ranking of features by attribution value in descending order. We progressively eliminate information from the top $k$ features and observe the trajectory of the model's output logit:
\begin{equation}
F^{(k)} = \frac{1}{N}\sum_{n=1}^{N} F_k\big(\tilde{\mathbf{x}}^{(n)}_{\pi_{1:k}}\big)
\end{equation}
where $\tilde{\mathbf{x}}^{(n)}_{\pi_{1:k}}$ represents the input after masking the top $k$ features. For masking in the spatial dimension, channel-wise mean 
replacement is employed, whereas Gaussian noise with 
matched mean and variance is used for the temporal dimension. The confidence curve $\{F^{(0)}, F^{(1)}, \ldots, F^{(K)}\}$ intuitively illustrates the causal validity of the attribution ranking: a steeper decline indicates that the attribution more precisely identifies the features critical to the decision.

Based on this, AOPC is used to quantify overall faithfulness:
\begin{equation}
\text{AOPC} = \frac{1}{K}\sum_{k=1}^{K}\big[F^{(0)} - F^{(k)}\big]
\end{equation}
Additionally, random-order masking serves as a baseline, with $\text{AOPC}_{\text{gain}} = \text{AOPC}_{\text{attr}} - \text{AOPC}_{\text{rand}}$ quantifying the gain of the attribution ranking relative to the random baseline.

AOPC evaluates the quality of the global ranking under cumulative masking. To complement this, the framework incorporates feature-wise independent intervention validation: independent single-feature masking is performed for each feature dimension (channel $c$ or temporal patch $p$) to calculate the actual logit drop $\Delta_i$:
\begin{equation}
    \Delta_i = \frac{1}{N} \sum_{n=1}^{N} \left[ f(x^{(n)}) - f(\tilde{x}^{(n)}_i) \right]
\end{equation}
where $\tilde{x}^{(n)}_i$ denotes sample $n$ with feature $i$ masked.

Spearman rank correlation is then computed between the attribution vector and the intervention impact vector:
\begin{equation}
\rho_{\text{spatial}} = r_s\big(\bar{\mathbf{s}},\; \boldsymbol{\Delta}\big), \quad
\rho_{\text{temporal}} = r_s\big(\bar{\mathbf{t}},\; \boldsymbol{\Delta}\big)
\end{equation}
where $\bar{\mathbf{s}}$ represents population-level channel importance and $\bar{\mathbf{t}}$ represents population-level temporal importance. A value of $\rho$ significantly greater than zero indicates monotonic consistency between the attribution ranking and the model's true feature dependencies. This metric enables cross-method comparisons, facilitating the recommendation of the most faithful attribution method for each model-task combination.

\subsubsection{Cross-Method Consistency}
For the same model-task combination, discrepancies may arise between the population attribution vectors produced by different attribution methods. This reflects the inherent assumption biases of the attribution methods themselves~\cite{krishnadisagreement}. The framework introduces cross-method consistency as a supplementary metric for attribution robustness by calculating the Spearman rank correlation between all pairs of methods:
\begin{equation}
\rho_{ij} = r_s\big(\bar{\mathbf{s}}^{(i)},\; \bar{\mathbf{s}}^{(j)}\big), \quad \bar{\rho} = \frac{2}{L(L-1)}\sum_{i<j}\rho_{ij}
\end{equation}
where $\bar{\mathbf{s}}^{(i)}$ is the population-level channel importance vector produced by the $i$-th attribution method, and $L$ is the number of attribution methods. A high consistency score $\bar{\rho}$ indicates that the attribution conclusions are insensitive to the choice of method, implying strong robustness of the results. In cases where discrepancies arise among different methods, the aforementioned AOPC and Spearman-based faithfulness metrics can serve as references to select the method with the best faithfulness as the recommended explanation for the specific model-task combination.

\subsection{LLM-assisted Clinical Interpretation}
The aforementioned analysis distills model decisions into low-dimensional attribution maps. However, as attribution patterns are merely numerical rankings, their neurophysiological significance requires interpretation through domain knowledge. To address this, the framework incorporates an LLM-assisted interpretation module. It leverages the medical and neuroscience knowledge internalized by the LLM to assess the physiological plausibility of attribution results and align them with clinical semantics. Attribution outputs are presented to the LLM as either structured data or visualizations, accompanied by a consistent set of domain-specific prompt constraints. These constraints require the LLM to perform reasoning within a neuroelectrophysiological framework, cross-referencing attribution patterns with the functional specialization of brain regions and the physiological significance of neural rhythms. Simultaneously, the LLM must integrate information across dimensions and assign confidence scores to uncertain inferences.

This design aims to bridge the semantic gap between numerical attribution and clinical understanding, allowing clinicians without computational backgrounds to comprehend the rationale behind model decisions. Simultaneously, physiological alignment assessments grounded in domain knowledge serve as supplementary evidence for the validity of the attributions. The output of this interpretive overlay does not influence fidelity assessments or alter attribution rankings. Instead, it functions as a downstream semantic interface within the attribution pipeline.

\section{Experimental Setup}

To validate the effectiveness of the proposed multidimensional EEG attribution analysis framework, we conducted systematic experiments on different models ,tasks and methods.

\textbf{Base Models.}
As shown in Table~\ref{tab:supported_models}, we selected five representative baseline EEG models for interpretability analysis, comprising four Transformer-based architectures (CBraMod~\cite{wang2025cbramod}, EEGPT, LaBraM, and BIOT) and one SSM-based architecture (EEGMamba). All models were fine-tuned on downstream tasks using their respective official pre-trained weights.

\textbf{Datasets and Tasks.}
As shown in Table~\ref{tab:dataset_summary}, we selected three datasets covering state recognition and event classification tasks:

(1) Temple University Hospital Abnormal EEG Corpus (TUAB)~\cite{obeid2016temple}: A binary classification task (normal vs. abnormal) using resting-state clinical EEG recordings across 2,993 sessions;
(2)Mumtaz2016~\cite{mumtaz2016mdd}:  A classification task distinguishing healthy controls (HC, $n=30$) from patients with major depressive disorder (MDD, $n=34$), utilizing 19-channel resting-state EEG data from 64 subjects in total;
(3) Temple University Hospital EEG Events Corpus (TUEV)~\cite{obeid2016temple}: A six-class classification task for fine-grained EEG micro-event detection, with annotations covering specific electrographic events such as spike-and-wave (SPSW) complexes, generalized periodic epileptiform discharges (GPEDs), and artifacts.

 We demonstrate the various processes of the framework using Mumtaz, while conducting quantitative analyses using TUAB and TUEV.
After undergoing model-specific preprocessing workflows, these datasets are used for model fine-tuning and inference.

\textbf{Attribution Methods.}
The framework integrates six post-hoc attribution methods (Integrated Gradients (IG)~\cite{sundararajan2017axiomatic}, SHAP~\cite{lundberg2017unified}, GradientSHAP~\cite{erion2021improving,lundberg2017unified}, LIME~\cite{ribeiro2016should}, Occlusion~\cite{zeiler2014visualizing}, and GradCAM~\cite{selvaraju2017grad})across three paradigms: input gradient-based, perturbation-based, and feature activation-based approaches.

\textbf{Evaluation Protocol.}
All analyses are based on population-level statistics. Specifically, evaluation metrics are calculated by averaging the attributions of true positive (TP) samples(defined as those with a model output probability of at least 0.7 for the target class). Spatial fidelity is assessed by computing Spearman's $\rho$ between attribution rankings and single-channel masking impact. Temporal fidelity is evaluated similarly by masking time segments sequentially. 
Frequency-band attribution is performed by attributing power within specific bands to generate joint spatial-frequency attribution maps. Consistency across different methods is quantified using the average 
Spearman rank correlation coefficient of the attribution rankings.

\begin{table}[htbp]
\centering
\caption{Overview of EEG foundation models supported by the proposed framework.}
\label{tab:supported_models}
\begin{tabular}{lll}
\toprule
\textbf{Model} & \textbf{Architecture} & \textbf{Pre-training Strategy} \\
\midrule
CBraMod  & Transformer & Masked EEG Modeling \\
LaBraM   & Transformer & Masked EEG Modeling \\
EEGPT    & Transformer & Autoregressive Pre-training \\
EEGMamba & SSM (Mamba) & Masked EEG Modeling \\
BIOT     & Transformer & Multi-dataset Pre-training \\
\bottomrule
\end{tabular}
\vspace{4pt}
\begin{minipage}{\linewidth}
\small \textit{Note:} The framework provides a unified adapter interface, supporting attribution analysis across heterogeneous architectures.
\end{minipage}
\end{table}

\begin{table*}[htbp]
\centering
\caption{Summary of the benchmark datasets used for explainability evaluation.}
\label{tab:dataset_summary}
\begin{tabular}{lm{2.5cm}p{8cm}l}
\toprule
\textbf{Dataset} & \textbf{Type} & \textbf{Description} & \textbf{Classes} \\
\midrule
Mumtaz2016 & State recognition  &  Binary classification of clinical EEG into major depressive disorder (MDD) vs. healthy control (HC) states & \parbox[t]{2.5cm}{0: HC \\ 1: MDD} \\[12pt]
\midrule
TUAB & State recognition & Binary classification of clinical EEG into normal vs.\ pathological states. & \parbox[t]{2.5cm}{0: Normal \\ 1: Abnormal} \\[12pt]
\midrule
TUEV & Event detection & Segment-level classification of clinical EEG into six neurological event types. & \parbox[t]{2.5cm}{0: SPSW 1: GPED \\ 2: PLED 3: EYEM \\ 4: ARTF  5: BCKG} \\[12pt]
\bottomrule
\end{tabular}

\vspace{4pt}
\begin{minipage}{\linewidth}
{\small \textit{Note:} SPSW = Spike and Sharp Wave; GPED = Generalized Periodic Epileptiform Discharges; PLED = Periodic Lateralized Epileptiform Discharges; EYEM = Eye Movement; ARTF = Artifact; BCKG = Background.}
\end{minipage}
\end{table*}

\begin{table*}[t]
\centering
\caption{Attribution method comparison on mumtaz2016 (MDD task), true positive samples. Boldface: channels in top-5 of $\geq$4/5 methods (excl.\ GradCAM). 
\label{tab:mumtaz_tp}
\underline{Underline}: clinically relevant channels (HC: posterior alpha --- O1/O2/Pz; MDD: frontal asymmetry --- F3/F4/F7/F8/Fp1/Fp2). $^\star$Recommended. $^{***}p<.001$, $^{**}p<.01$, $^{*}p<.05$.}
\label{tab:mumtaz_tp}
\scriptsize
\resizebox{\textwidth}{!}{
\begin{tabular}{llccccccc}
\toprule
 & & \multicolumn{3}{c}{Class 0 — HC (N=100)} & \multicolumn{3}{c}{Class 1 — MDD (N=100)} \\
\cmidrule(lr){3-5} \cmidrule(lr){6-8}
Model & Method & $\rho$ & Consist. & Top-5 Channels & $\rho$ & Consist. & Top-5 Channels \\
\midrule
\multirow{6}{*}{\shortstack[l]{CBraMod\\(Transformer)}}
 & IG$^\star$        & $0.981^{***}$ & 0.712 & \underline{\textbf{O1, O2, Pz}}, T3, T6  & $0.972^{***}$  & 0.902 & C3, \textbf{C4}, \underline{\textbf{F3}}, \underline{F4}, \underline{\textbf{F8}} \\
 & GradientSHAP      & $0.979^{***}$ & 0.709 & \underline{\textbf{O1, O2, Pz}}, T3, T6  & $0.968^{***\star}$ & 0.907 & C3, \textbf{C4}, \underline{\textbf{F3}}, \underline{F4}, \underline{\textbf{F8}} \\
 & SHAP              & $0.961^{***}$ & 0.702 & \underline{\textbf{O1, O2, Pz}}, T3, T6  & $0.933^{***}$  & 0.901 & \textbf{C4}, Cz, \underline{\textbf{F3}}, \underline{F4}, \underline{\textbf{F8}} \\
 & LIME              & $0.953^{***}$ & 0.695 & \underline{\textbf{O1, O2, Pz}}, T3, T6  & $0.949^{***}$  & 0.874 & \textbf{C4}, \underline{\textbf{F3}}, \underline{\textbf{F8}}, \underline{Fp1}, \underline{Fp2} \\
 & Occlusion         & $0.947^{***}$ & 0.700 & \underline{\textbf{O1, O2}}, T3, T4, T6  & $0.877^{***}$  & 0.839 & \textbf{C4}, Cz, \underline{\textbf{F3}}, \underline{Fp1}, \underline{Fp2} \\
 & GradCAM           & $-0.339$      & $-$0.355 & C3, F4, F8, \underline{Pz}, T3  & $0.702^{***}$  & 0.728 & \textbf{C4}, Cz, \underline{\textbf{F3}}, \underline{F4}, Fz \\
\midrule
\multirow{6}{*}{\shortstack[l]{EEGMamba\\(SSM)}}
 & IG                & $0.532^{*}$   & 0.289 & F7, Fp1, Fp2, T3, T4  & $0.884^{***}$  & 0.756 & \textbf{C4}, \underline{F3}, \underline{\textbf{F8}}, \underline{\textbf{Fp1}}, \textbf{T5} \\
 & GradientSHAP      & $0.793^{***}$ & 0.576 & \textbf{C4}, F7, \textbf{F8}, \underline{\textbf{O1}}, \underline{\textbf{O2}}  & $0.919^{***}$  & 0.795 & \textbf{C4}, \underline{\textbf{F8}}, \underline{\textbf{Fp1}}, \underline{Fp2}, \textbf{T5} \\
 & SHAP              & $0.802^{***}$ & 0.559 & \textbf{C4}, \textbf{F8}, Fp1, \underline{\textbf{O1}}, \underline{\textbf{O2}}  & $0.946^{***}$  & 0.801 & \textbf{C4}, \underline{\textbf{F8}}, \underline{\textbf{Fp1}}, \underline{Fp2}, \textbf{T5} \\
 & LIME              & $0.844^{***}$ & 0.584 & \textbf{C4}, \textbf{F8}, \underline{\textbf{O1}}, \underline{\textbf{O2}}, \underline{Pz}  & $0.930^{***\star}$ & 0.806 & \textbf{C4}, \underline{\textbf{F8}}, \underline{\textbf{Fp1}}, P4, \textbf{T5} \\
 & Occlusion$^\star$ & $0.911^{***}$ & 0.568 & \textbf{C4}, \textbf{F8}, \underline{\textbf{O1}}, \underline{\textbf{O2}}, \underline{Pz}  & $0.865^{***}$  & 0.757 & \textbf{C4}, Cz, \underline{\textbf{F8}}, \underline{\textbf{Fp1}}, P4 \\
 & GradCAM           & $-0.332$      & $-$0.266 & C3, F3, F4, Fz, \underline{Pz}  & $0.204$        & 0.216 & Cz, P3, P4, Pz, T4 \\
\bottomrule
\end{tabular}
}
\end{table*}

\begin{table*}[htbp]
\centering
\caption{Channel-level explainability comparison on TUAB (Class~0: subject\_aaaaaiwz, Normal; Class~1: subject\_aaaaamft, Abnormal).
\textbf{Boldface}: channels in top-5 of $\geq$4/5 methods (excl.\ GradCAM).
\label{tab:tuab_explainability}
\underline{Underline}: clinically relevant channels (Normal: posterior alpha --- O1/O2, P3/P4; Abnormal: temporal focus --- T1/T2/T3/T4/T5/T6, F7/F8).
$\star$~Recommended. $^{***}p<.001$, $^{**}p<.01$, $^{*}p<.05$.}
\setlength{\tabcolsep}{3pt}
\fontsize{6.5pt}{7.8pt}\selectfont
\begin{tabular}{ll ccl ccl}
\toprule
& & \multicolumn{3}{c}{Class 0 --- Normal} & \multicolumn{3}{c}{Class 1 --- Abnormal} \\
\cmidrule(lr){3-5} \cmidrule(lr){6-8}
Model & Method & $\rho$ & Consist. & Top-5 Channels & $\rho$ & Consist. & Top-5 Channels \\
\midrule
\multirow{6}{*}{\shortstack[l]{CBraMod\\{\scriptsize(116/272)}}}
& IG$^\star$       & $0.712^{**}$  & 0.643 & \underline{\textbf{C3-P3}}, FP2-F4, \underline{\textbf{P4-O2}}, \textbf{T3-T5}, \underline{T6-O2} & $0.979^{***}$ & 0.872 & \underline{\textbf{F8-T4}}, \textbf{FP2-F8}, \underline{\textbf{T3-T5}}, \underline{\textbf{T5-O1}}, \underline{\textbf{T6-O2}} \\
& SHAP             & $0.691^{**}$  & 0.639 & \underline{\textbf{C3-P3}}, F3-C3, F7-T3, FP2-F8, \underline{\textbf{P4-O2}}          & $0.956^{***}$ & 0.872 & \underline{\textbf{F8-T4}}, \textbf{FP2-F8}, P3-O1, \underline{\textbf{T5-O1}}, \underline{\textbf{T6-O2}} \\
& GradientShap     & $0.712^{**}$  & 0.641 & \underline{\textbf{C3-P3}}, FP2-F4, \underline{\textbf{P4-O2}}, \textbf{T3-T5}, \underline{T6-O2} & $0.979^{***}$ & 0.872 & \underline{\textbf{F8-T4}}, \textbf{FP2-F8}, \underline{\textbf{T3-T5}}, \underline{\textbf{T5-O1}}, \underline{\textbf{T6-O2}} \\
& LIME             & 0.279         & 0.431 & F7-T3, FP2-F4, \underline{\textbf{P4-O2}}, \textbf{T3-T5}, \underline{T5-O1}          & $0.979^{***}$ & 0.872 & \underline{\textbf{F8-T4}}, \textbf{FP2-F8}, \underline{\textbf{T3-T5}}, \underline{\textbf{T5-O1}}, \underline{\textbf{T6-O2}} \\
& Occlusion        & $0.594^{*}$   & 0.562 & \underline{\textbf{C3-P3}}, FP2-F8, \underline{\textbf{P4-O2}}, \textbf{T3-T5}, T4-T6 & $0.991^{***}$ & 0.872 & \underline{\textbf{F8-T4}}, \textbf{FP2-F8}, \underline{\textbf{T3-T5}}, \underline{\textbf{T5-O1}}, \underline{\textbf{T6-O2}} \\
& GradCAM          & 0.121         & $-$0.013 & \underline{C3-P3}, \underline{C4-P4}, F4-C4, T3-T5, \underline{T5-O1}                          & 0.465         & 0.475 & \underline{F8-T4}, FP2-F8, P3-O1, \underline{T4-T6}, \underline{T6-O2} \\
\midrule
\multirow{6}{*}{\shortstack[l]{EEGMamba\\{\scriptsize(116/270)}}}
& LIME$^\star$     & $0.976^{***}$ & 0.714 & \textbf{Fp1-F3}, \textbf{Fp2-F4}, \underline{P3-O1}, \textbf{T3-T5}, \textbf{T4-T6} & $0.738^{**}$  & 0.516 & C4-P4, F3-C3, \underline{F8-T4}, Fp1-F7, P4-O2 \\
& IG               & $0.591^{*}$   & 0.493 & F8-T4, \textbf{Fp1-F3}, \textbf{T3-T5}, \textbf{T4-T6}, \underline{T6-O2} & $0.712^{**}$  & 0.542 & \underline{F8-T4}, Fp1-F3, P3-O1, P4-O2, \underline{T4-T6} \\
& SHAP             & $0.856^{***}$ & 0.678 & \textbf{Fp1-F3}, \textbf{Fp2-F4}, \underline{P4-O2}, \textbf{T3-T5}, \textbf{T4-T6} & $0.859^{***}$ & 0.632 & F3-C3, F4-C4, \underline{F7-T3}, \underline{F8-T4}, Fp2-F4 \\
& GradientShap     & $0.832^{***}$ & 0.654 & F3-C3, \textbf{Fp1-F3}, \textbf{Fp2-F4}, \textbf{T3-T5}, \textbf{T4-T6} & $0.868^{***}$ & 0.671 & \underline{F8-T4}, Fp1-F3, Fp2-F8, P4-O2, \underline{T6-O2} \\
& Occlusion        & $0.991^{***}$ & 0.690 & F8-T4, \textbf{Fp2-F4}, \underline{P3-O1}, \textbf{T3-T5}, \textbf{T4-T6} & $0.900^{***}$ & 0.542 & \underline{F7-T3}, \underline{F8-T4}, Fp1-F3, Fp1-F7, Fp2-F8 \\
& GradCAM          & 0.238         & 0.038 & F3-C3, F7-T3, F8-T4, Fp1-F7, \textbf{T4-T6}                   & 0.185         & 0.115 & \underline{F8-T4}, Fp1-F3, Fp1-F7, Fp2-F4, \underline{T4-T6} \\
\midrule
\multirow{6}{*}{\shortstack[l]{LaBraM\\{\scriptsize(116/272)}}}
& Occlusion$^\star$ & $0.720^{***}$ & 0.359 & C4, F3, F4, FZ, \textbf{T2}                                  & $0.627^{**}$  & 0.238 & CZ, F4, FP1, P3, \underline{T2} \\
& IG               & $-$0.146      & 0.152 & A2, C4, \underline{O2}, \underline{P3}, T6                    & $0.551^{**}$  & 0.176 & F3, F4, \underline{F7}, P3, \underline{T5} \\
& SHAP             & 0.231         & 0.187 & FP1, \underline{O2}, T1, \textbf{T2}, T3                      & 0.360         & 0.202 & A1, A2, F3, \underline{F8}, PZ \\
& GradientShap     & 0.325         & 0.345 & A2, \underline{O2}, \underline{P4}, \textbf{T2}, T6           & 0.316         & 0.314 & A1, A2, \underline{T3}, \underline{T4}, \underline{T5} \\
& LIME             & 0.027         & 0.371 & FZ, \underline{P3}, \underline{P4}, \textbf{T2}, T3           & 0.223         & 0.173 & A1, CZ, F4, FP2, \underline{T3} \\
& GradCAM          & 0.384         & 0.102 & A2, FP1, FP2, \underline{P4}, \textbf{T2}                    & 0.256         & 0.254 & \underline{F7}, FP1, P4, \underline{T3}, \underline{T4} \\
\midrule
\multirow{6}{*}{\shortstack[l]{EEGPT\\{\scriptsize(115/264)}}}
& Occlusion$^\star$ & $0.717^{***}$ & 0.312 & \textbf{A2}, CZ, FP1, T3, T6                       & $0.670^{***}$ & 0.125 & \underline{F8}, \textbf{\underline{T2}}, \underline{T3}, \textbf{\underline{T4}}, \underline{T6} \\
& GradientShap     & $0.628^{**}$  & 0.274 & \textbf{A2}, F8, FP1, \underline{O2}, T5 & $0.589^{**}$  & 0.561 & A2, O1, O2, \textbf{\underline{T2}}, \textbf{\underline{T4}} \\
& LIME             & 0.297         & 0.407 & \textbf{A2}, C3, \underline{O1}, \underline{O2}, T6             & $0.656^{***}$ & 0.420 & CZ, O1, O2, \textbf{\underline{T2}}, \textbf{\underline{T4}} \\
& IG               & 0.058         & 0.245 & \textbf{A2}, \underline{O1}, \underline{O2}, T3, T4 & $0.490^{*}$   & 0.500 & A2, O1, O2, \textbf{\underline{T2}}, \textbf{\underline{T4}} \\
& SHAP             & 0.008         & 0.226 & \textbf{A2}, F7, \underline{O1}, T3, T6             & 0.226         & 0.299 & FP1, \underline{T3}, \textbf{\underline{T4}}, \underline{T5}, \underline{T6} \\
& GradCAM          & ---           & ---   & F3, F4, F7, FP2, T2                                            & ---           & ---   & F3, F4, \underline{F7}, FP2, \underline{T2} \\
\midrule
\multirow{6}{*}{\shortstack[l]{BIOT\\{\scriptsize(75/270)}}}
& Occlusion$^\star$ & $-$0.412      & 0.285 & \underline{C3-P3}, \underline{C4-P4}, F3-C3, F8-T4, T4-T6     & $0.762^{***}$ & 0.331 & \underline{\textbf{P4-O2}}, \underline{T3-T5}, \underline{T4-T6}, \underline{\textbf{T5-O1}}, \underline{\textbf{T6-O2}} \\
& GradCAM          & $0.653^{**}$  & ---   & \underline{C4-P4}, \underline{P3-O1}, \underline{P4-O2}, \underline{T5-O1}, \underline{T6-O2} & 0.450         & ---   & \underline{F7-T3}, \textbf{FP1-F3}, FP1-F7, FP2-F8, \underline{\textbf{T5-O1}} \\
& LIME             & 0.235         & 0.358 & F3-C3, \textbf{FP1-F7}, \underline{P3-O1}, \underline{T3-T5}, \underline{T6-O2} & 0.268         & 0.642 & \underline{C3-P3}, \textbf{FP1-F3}, \underline{\textbf{P4-O2}}, \underline{T3-T5}, \underline{\textbf{T6-O2}} \\
& IG               & $-$0.350      & 0.490 & \underline{C4-P4}, FP1-F3, \textbf{FP1-F7}, \underline{P3-O1}, \underline{P4-O2} & 0.094         & 0.624 & C3-P3, \textbf{FP1-F3}, \underline{\textbf{P4-O2}}, \underline{\textbf{T5-O1}}, \underline{\textbf{T6-O2}} \\
& GradientShap     & $-$0.374      & 0.418 & \underline{C4-P4}, FP1-F3, \textbf{FP1-F7}, \underline{P4-O2}, \underline{T6-O2} & $-$0.021      & 0.675 & F4-C4, \textbf{FP1-F3}, \underline{P3-O1}, \underline{\textbf{P4-O2}}, \underline{\textbf{T5-O1}} \\
& SHAP             & $-$0.235      & 0.401 & \underline{C3-P3}, \textbf{FP1-F7}, \underline{P3-O1}, T4-T6, \underline{T6-O2} & $-$0.018      & 0.586 & \textbf{FP1-F3}, \underline{\textbf{P4-O2}}, \underline{T4-T6}, \underline{\textbf{T5-O1}}, \underline{\textbf{T6-O2}} \\
\bottomrule
\end{tabular}
\end{table*}

\begin{table}[htbp]
\centering
\caption{Temporal faithfulness on TUEV Class~1 (GPED). $\rho$: Spearman correlation between patch attribution rank and confidence drop rank.
$\star$~Recommended (highest $\rho>0$ with $p<.05$). $^{***}p<.001$, $^{**}p<.01$, $^{*}p<.05$. ---: not computable (constant attribution).
\textit{Note:} The patch granularities for the models are: CBraMod / EEGMamba / LaBraM = 5, BIOT = 9, and EEGPT = 15. The absolute magnitudes of $\rho$ are not directly comparable; a larger $n$ imposes stricter requirements on the method's event localization accuracy but simultaneously results in a lower significance threshold.}
\label{tab:tuev_temporal_faithfulness}
\setlength{\tabcolsep}{4pt}
\fontsize{6.5pt}{7.8pt}\selectfont
\begin{tabular}{ll c}
\toprule
Model & Method & $\rho$ \\
\midrule
\multirow{6}{*}{CBraMod}
& LIME             & 0.800 \\
& IG               & 0.800 \\
& SHAP             & 0.800 \\
& GradientSHAP     & 0.800 \\
& Occlusion        & $-0.700$ \\
& GradCAM          & $-0.900^{*}$ \\
\midrule
\multirow{6}{*}{EEGMamba}
& LIME             & $-0.300$ \\
& IG               & 0.600 \\
& SHAP$^\star$     & $1.000^{***}$ \\
& GradientSHAP     & 0.800 \\
& Occlusion        & $-0.400$ \\
& GradCAM          & 0.300 \\
\midrule
\multirow{6}{*}{EEGPT}
& LIME             & 0.132 \\
& IG               & 0.064 \\
& SHAP             & 0.204 \\
& GradientSHAP     & $0.543^{*}$ \\
& Occlusion$^\star$ & $0.686^{**}$ \\
& GradCAM          & --- \\
\midrule
\multirow{6}{*}{LaBraM}
& LIME             & 0.100 \\
& IG               & 0.300 \\
& SHAP             & 0.100 \\
& GradientSHAP     & 0.700 \\
& Occlusion        & 0.200 \\
& GradCAM          & $-0.100$ \\
\midrule
\multirow{6}{*}{BIOT}
& LIME             & 0.771 \\
& IG$^\star$       & $0.700^{*}$ \\
& SHAP             & 0.029 \\
& GradientSHAP     & $0.636^{*}$ \\
& Occlusion        & 0.657 \\
& GradCAM          & 0.536 \\
\bottomrule
\end{tabular}
\end{table}

\begin{figure*}[htbp]
\centering
\subfloat[Class 0 (Control) -- Channel Attribution]{%
  \includegraphics[width=0.5\linewidth]{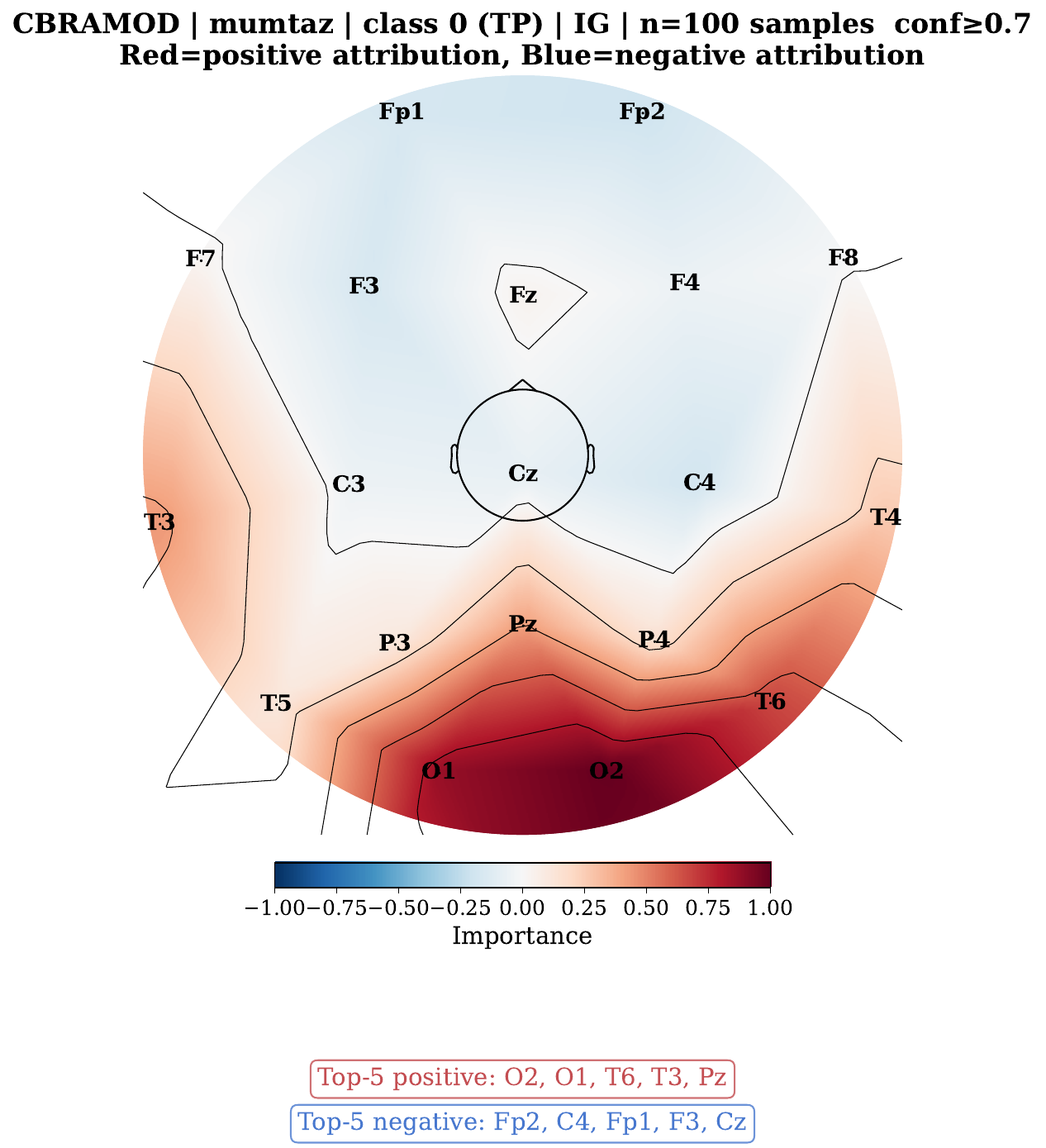}}%
\hfill
\subfloat[Class 1 (MDD) -- Channel Attribution]{%
  \includegraphics[width=0.5\linewidth]{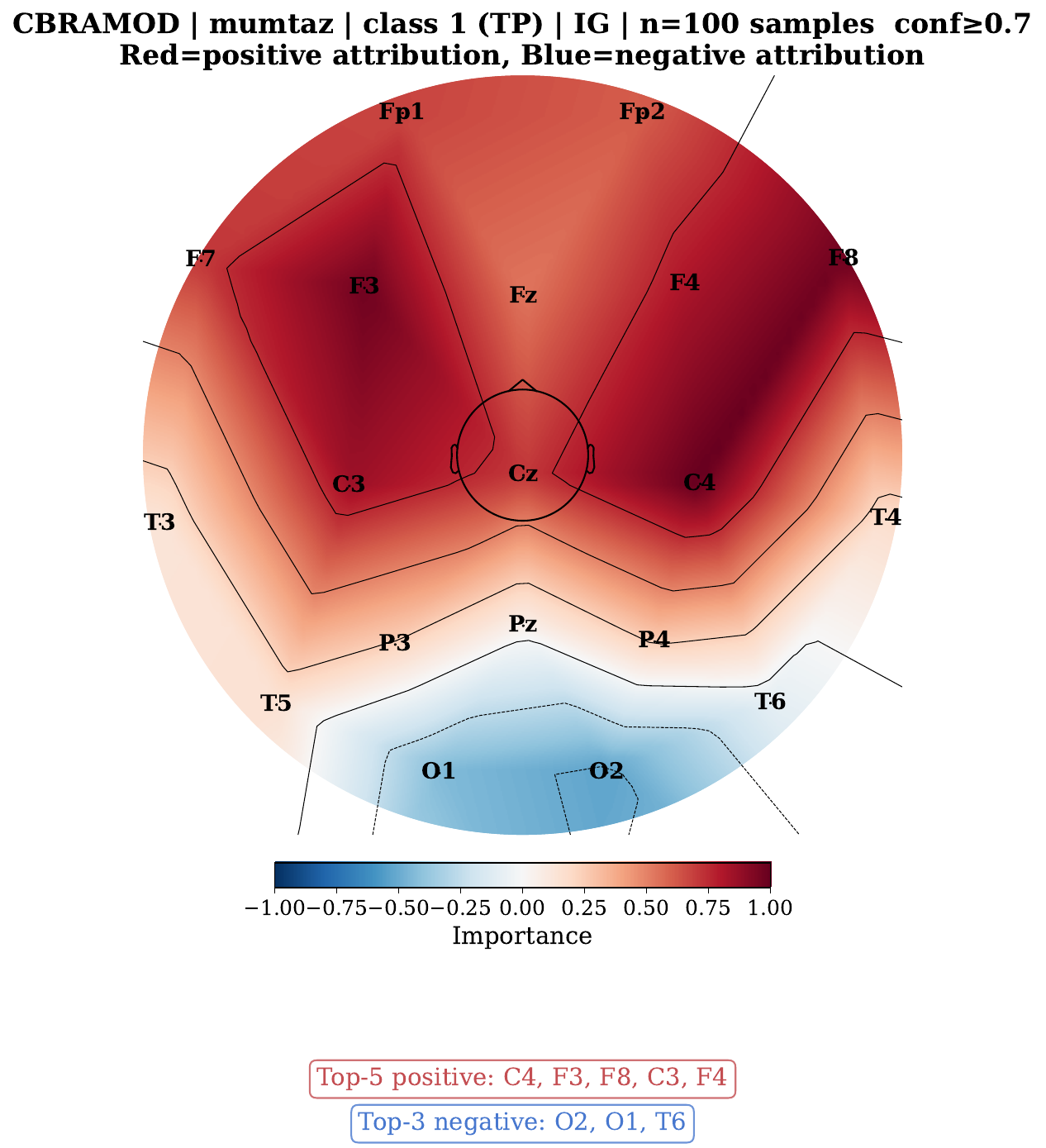}}

\vspace{4pt}

\subfloat[Class 0 (Control) -- Spatial Faithfulness]{%
  \includegraphics[width=0.45\linewidth]{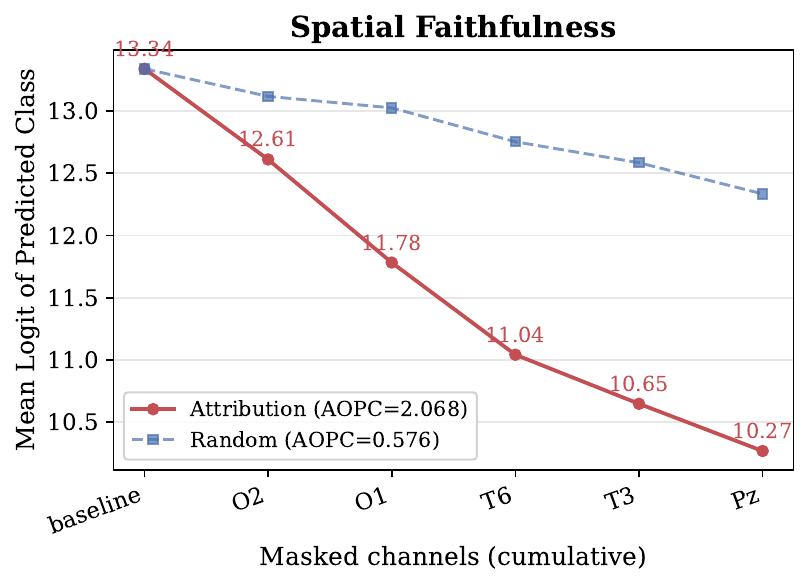}}%
\hfill
\subfloat[Class 1 (MDD) -- Spatial Faithfulness]{%
  \includegraphics[width=0.45\linewidth]{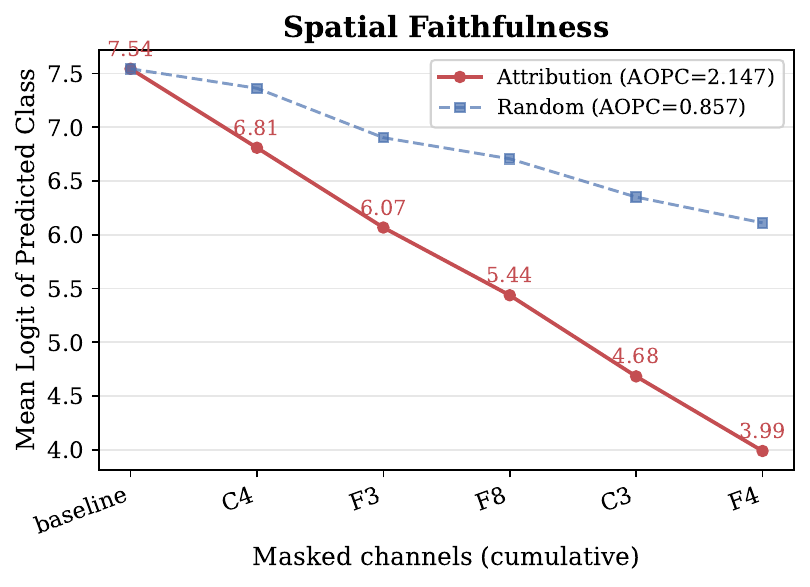}}
\caption{Channel-level attribution analysis on the Mumtaz dataset using IG with CBraMod. Top row: topographic maps of channel importance. Bottom row: spatial faithfulness evaluation.}
\label{fig:mumtaz_ig_channel}
\end{figure*}
\begin{figure*}[htbp]
\centering
\subfloat[Class 0 (Control)]{
    \includegraphics[width=1.0\textwidth]{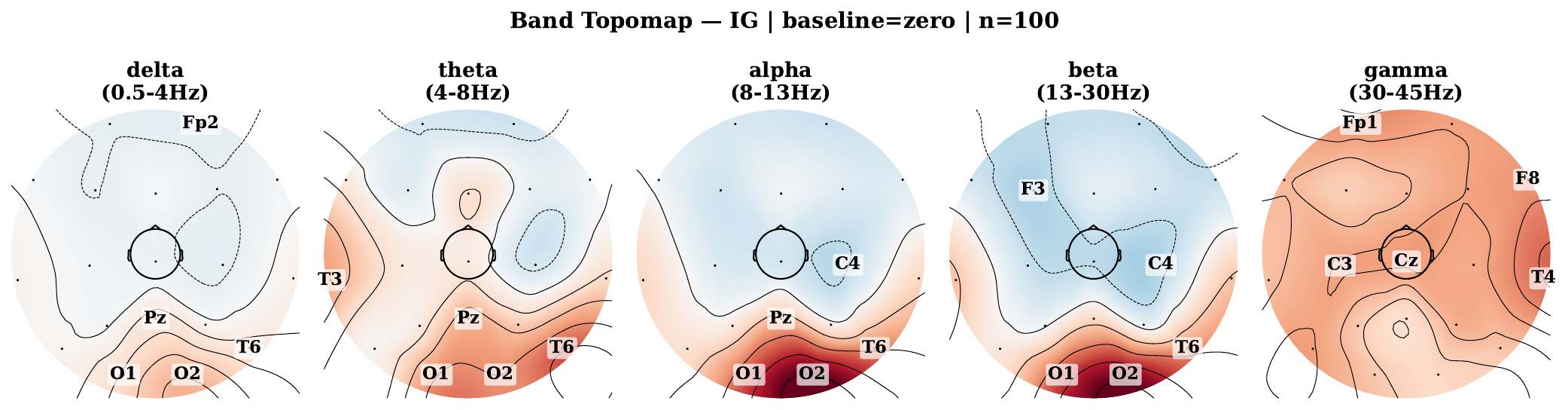}}

\vspace{-0.1cm}

\subfloat[Class 1 (MDD)]{
    \includegraphics[width=1.0\textwidth]{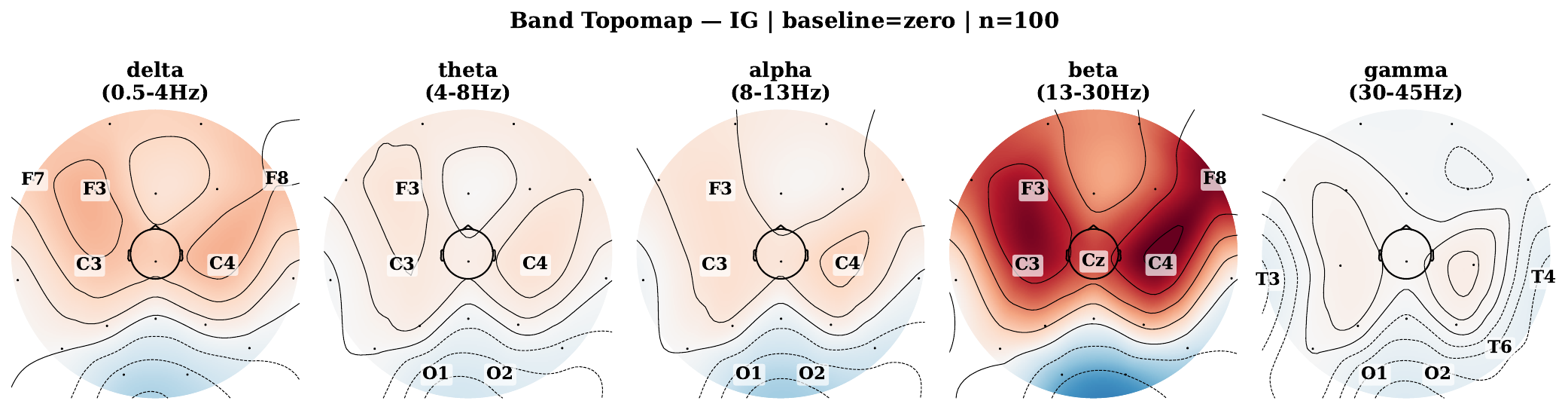}}
\caption{Band-level population attribution on the Mumtaz dataset using IG with CBraMod. Each row shows the topographic distribution of attribution across five frequency bands for one class.}
\label{fig:mumtaz_ig_band}
\end{figure*}

\subsection{Resting-State Task: Spatial and Frequency-Band Attribution Analysis}

\subsubsection{Channel-Level Spatial Fidelity}

Tables ~\ref{tab:mumtaz_tp} and ~\ref{tab:tuab_explainability} present the channel-wise spatial fidelity of various \textbf{model-method} combinations on the Mumtaz and TUAB datasets, respectively.

\textbf{Mumtaz Dataset.}
On the Mumtaz dataset (Table~\ref{tab:mumtaz_tp}), CBraMod’s attributions for the HC class consistently pointed to the occipital regions O1, O2, and Pz (IG: $\rho=0.981$, $p<.001$), while for the MDD class, they focused on the frontal regions F3 and F8 and the central region C4. Figure~\ref{fig:mumtaz_ig_channel} displays the group-averaged attribution topographies and faithfulness curve for the two categories using the IG method. Differences in lateralization are evident between the groups: the HC (Healthy Control) group exhibits strong positive attribution in the occipital region, whereas the MDD (Major Depressive Disorder) group shows strong positive attribution in the fronto-central region.The attribution-guided curves show a steeper decline than random baselines across both classes, which indicates that the attribution 
ranking effectively captures feature importance patterns aligned 
with the model's decision process.

From the perspective of clinical interpretability, the occipital attribution pattern observed in the healthy control (HC) group aligns with the physiological characteristics of resting-state EEG in healthy individuals, where the occipital lobe serves as a core region of activation ~\cite{niedermeyer2005electroencephalography}. Meanwhile, the prominent frontal lobe attribution features in the major depressive disorder (MDD) group are consistent with previous findings regarding abnormal activity and impaired neural function in the frontal brain regions of patients with depression~\cite{pizzagalli2011frontocingulate}.
EEGMamba also demonstrated good spatial fidelity on the Mumtaz dataset (HC class Occlusion: $\rho=0.911$; MDD class LIME: $\rho=0.930$). Furthermore, its top-5 channels for the MDD class included frontal-temporal channels such as F8, Fp1, and T5, further corroborating the clinical consistency of the attributions across models.

\textbf{TUAB Dataset.}
On the TUAB dataset (Table~\ref{tab:tuab_explainability}), CBraMod and EEGMamba demonstrate high attribution fidelity. Taking CBraMod as an example, both IG and GradientSHAP achieve $\rho=0.979$ ($p<.001$) for the abnormal class. The top-5 channels (F8-T4, FP2-F8, T3-T5, T5-O1, T6-O2) are highly concentrated in the temporal region, consistent with the temporal lobe abnormal discharge patterns characteristic of abnormal EEG~\cite{tatum2021clinical}. The top-5 channels identified by five of the methods (excluding GradCAM) are identical, indicating high cross-method consistency in attribution results. For the normal class, the models focus on channels in the central-parieto-occipital regions (C3-P3, P4-O2), aligning with the dominance of posterior alpha rhythms in normal resting-state EEG~\cite{umemoto2021resting}.

In contrast, LaBraM and BIOT generally exhibited lower spatial fidelity (with most $\rho$ values failing to reach statistical significance). This may be linked to differences in downstream task performance on TUAB, as attribution fidelity presupposes that the model has actually learned meaningful spatial patterns. Among all models, GradCAM performed the worst. It stems from the fact that GradCAM was originally designed for CNN spatial feature maps and lacks a direct spatial correspondence when applied to Transformer architectures based on patch embeddings.

\subsubsection{Frequency-Band-Level Attribution Analysis}

Figure~\ref{fig:mumtaz_ig_band} displays the joint spatial-frequency attribution topographic maps for the CBraMod model on the Mumtaz dataset, derived using Integrated Gradients (IG). This analysis, conducted by applying IG to the power features of five standard frequency bands, reveals the model's reliance on specific frequency bands across different brain regions.

\textbf{HC Class} (Figure~\ref{fig:mumtaz_ig_band}, top): For healthy controls (Class 0), discriminative contributions in the alpha (8–13 Hz) and low-beta bands were concentrated in occipito-parietal regions, with relatively negative patterns in frontal areas. This aligns with the posterior alpha dominance typically observed in healthy subjects at rest. It is also consistent with MDD-related patterns, particularly the attenuation of posterior alpha activity and alterations in frontal midline slow-waves, both of which have been repeatedly validated in the Mumtaz dataset ~\cite{movahed2021major}. The gamma (30–45 Hz) band exhibited diffuse and relatively high-intensity attribution. Given that scalp gamma activity is susceptible to interference from temporalis/frontalis electromyographic (EMG) signals and microsaccade-locked components under standard preprocessing conditions, the signal likely contains non-neural components.

\textbf{MDD Class} (Figure~\ref{fig:mumtaz_ig_band}, bottom): Patients with depression (Class 1) exhibited positive contributions from fronto-temporal regions in the $\delta/\theta$ bands and formed significant hotspots in the fronto-central region within the $\beta$ band. These findings align with conclusions from recent machine learning-based EEG meta-analyses of major depressive disorder (MDD), which identified ``elevated frontal slow-wave power'' and ``altered $\beta$-band powe'' as robust and discriminative physiological markers~\cite{watts2022predicting,khosla2022automated}. The attribution maps showed near-symmetrical patterns for frontal $\alpha$ activity, without explicitly displaying the Frontal Alpha Asymmetry (FAA) pattern ~\cite{watts2022predicting}. This phenomenon suggests that the model relies on distributed power combinations across electrodes and frequency bands for classification. It likely implicitly integrates information related to FAA, rather than highlighting it as a distinct asymmetry metric at the attribution level.

Overall, both types of IG attribution across dominant frequency bands and spatial distributions align with mainstream findings from the Mumtaz dataset, validating the physiological interpretability of the model's decision.

The value of frequency-band-level attribution analysis lies in its ability to go beyond simple channel ranking and reveal the spectral basis underlying the model's decisions. While channel-level analysis uncovers \textbf{where} the model focuses, frequency-band-level analysis further elucidates \textbf{what} the model specifically perceives, thereby establishing a correspondence between the internal representations of ``black-box'' models and specific electrophysiological mechanisms. 

\subsection{Event-based Tasks: Temporal Faithfulness Analysis}
In the Table ~\ref{tab:tuev_temporal_faithfulness}, $\rho$ denotes the Spearman correlation coefficient between the ranking of segment attributions and the ranking of confidence drops following occlusion. A positive value close to 1 indicates that the temporal segments the model actually relies on align well with the attribution annotations. Since the number of patches ($n=5/9/15$) varies across models, the absolute values of $\rho$ cannot be directly compared. Larger values of $n$ imply finer granularity in event localization and a relatively lower significance threshold. Therefore, the conclusions are drawn based on statistical significance.

At finer granularity, EEGPT ($n=15$) achieved $\rho=0.686^{**}$ with Occlusion and $\rho=0.543^{*}$ with GradientSHAP, demonstrating strong evidence of event anchoring. BIOT ($n=9$) showed consistently positive values across all six methods, with IG and GradientSHAP reaching significance, indicating clear cross-method convergence.

CBraMod ($n=5$) consistently yielded $\rho=0.800$ across LIME, IG, SHAP, and GradientSHAP. While it did not reach statistical significance due to sample size limitations,this cross-method consistency offers moderate evidence for attribution reliability. Krishnadisagreement et al.~\cite{krishnadisagreement}, through a systematic comparison of various attribution methods, point out that in practice, consistency among different methods serves as a common heuristic for assessing the reliability of attributions. 

For EEGMamba ($n=5$), SHAP achieved $\rho=1.000^{***}$, whereas LIME and Occlusion showed inverse correlations, illustrating a typical case of inter-method disagreement~\cite{krishnadisagreement}. LaBraM ($n=5$) exhibited low and non-significant $\rho$ values across methods, reflecting overall weakness in patch-level temporal faithfulness. Bjelogrlic et al.\cite{turbe2023evaluation} analyzed similar low-faithfulness phenomena in XAI benchmarks for time-series classification, identifying common causes such as the use of distributed representations with cross-segment redundancy, high feature correlation leading to contextual compensation of perturbations, and method-architecture coupling. 

Across the different models, GradientSHAP consistently maintained a positive $\rho$ and frequently approached the best-performing results for each model, demonstrating relative robustness in the current evaluation. While SHAP exhibited outstanding local performance, it showed high variance across models. LIME and Occlusion showed inconsistent directional behavior across different models, reflecting the limited stability of these methods when applied to high-dimensional, highly autocorrelated signals such as EEG data.

\begin{figure*}[htbp]
\centering
\vspace{-10pt}      
\includegraphics[width=0.9\textwidth]{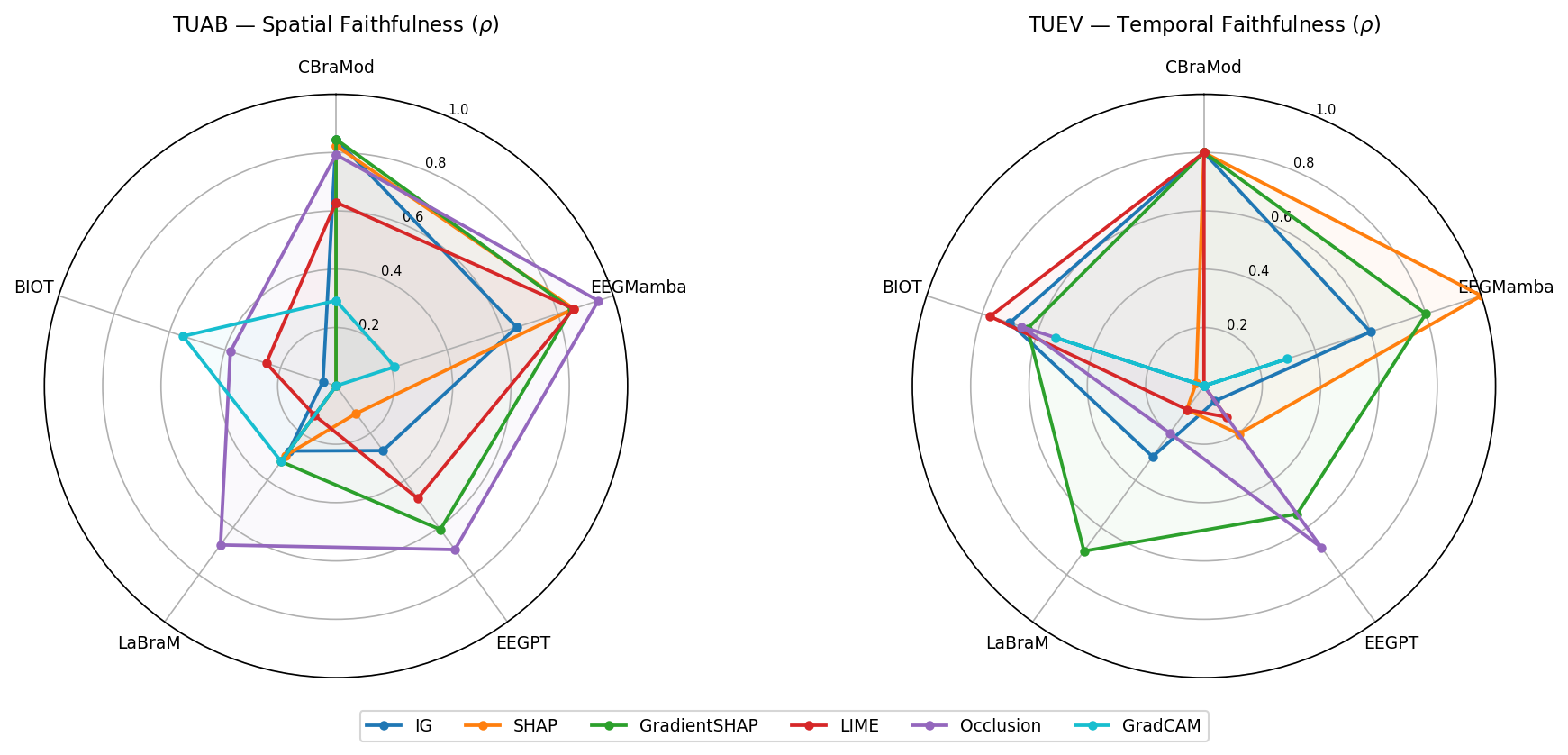}
\caption{Radar chart of attribution method faithfulness across models. Left: spatial faithfulness ($\rho$) on TUAB (averaged over Normal/Abnormal classes). Right: temporal faithfulness ($\rho$) on TUEV Class~1 (GPED). Negative $\rho$ values are clipped to 0.}
\label{fig:radar_chart}
\end{figure*}

\subsection{Discussion}

Experimental results validate the effectiveness of the proposed framework from multiple dimensions:

\textbf{Clinical consistency of attribution results.}
In the spatial dimension, various attribution methods consistently highlight brain regions that align with clinical knowledge: the TUAB abnormality category corresponds to temporal lobe regions, the Mumtaz HC (Healthy Control) category to posterior occipital regions, and the MDD (Major Depressive Disorder) category to prefrontal regions. Frequency band analysis further reveals the electrophysiological basis of model decisions, thereby establishing an interpretable link between ``black-box'' predictions and physiological mechanisms.

\textbf{Necessity of multi-method cross-validation.}
The radar chart (Fig. \ref{fig:radar_chart}) clearly demonstrates that no single attribution method performs optimally across all model-task combinations. For instance, the Occlusion method achieves a high correlation of $\rho=0.991$ on EEGMamba (TUAB) but only $-0.412$ on BIOT; the IG method performs stably on CBraMod but fails almost completely on BIOT (TUAB). These results justify the framework's design choice to integrate multiple methods and provide a recommendation mechanism.

\textbf{Spatial fidelity generally exceeds temporal fidelity.}
Overall, the radar radii for TUAB are significantly larger than those for TUEV, indicating that current EEG foundation models exhibit higher consistency in event anchoring along the channel/spatial dimension than along the temporal dimension. Attribution fidelity in the temporal dimension is generally low across most model-method combinations, reflecting that the stable characterization of patch-level temporal fidelity requires further fine-grained methodological evaluation.

\section{Conclusion}

This paper proposes a unified ex post-hoc interpretability framework for EEG basic models. This framework achieves multi-dimensional attribution analysis independent of specific model architectures by abstracting different model architectures and integrating various attribution methods.

Through population-level experiments on EEG tasks, we found that: (1) there is no single attribution method that is optimal across models and dimensions; method selection is highly coupled with model architecture and task dimensions; (2) the attribution consistency of current EEG basic models is generally higher in the spatial dimension than in the temporal dimension, and a more granular methodological evaluation is needed to stably characterize patch-level temporal fidelity; (3) cross-model comparisons for specific tasks reveal both common neurobiomarkers and model-specific biases.

\bibliographystyle{IEEEtran}
\bibliography{mybib.bib}

\end{document}